Table of Contents

# When Does Consensus Beat Voting? A Critical Analysis of Statistical Label Fusion in Medical Image Segmentation

Renjie He

Department of Radiation Oncology, The University of Texas MD Anderson Cancer Center, Houston, TX 77030, USA

**e-mail: mdacc2014@gmail.com*

**Purpose:** Consensus segmentation—combining multiple expert contours into a single reference standard—is foundational to radiation oncology practice, with STAPLE (Simultaneous Truth and Performance Level Estimation) as the dominant algorithm for two decades. Despite widespread adoption by NRG Oncology for contouring atlases and by multi-observer studies such as C3RO, the conditions under which STAPLE outperforms simple majority voting remain poorly understood. This paper provides a rigorous mathematical and empirical analysis of when consensus methods genuinely improve upon voting, and characterizes the failure modes most relevant to clinical practice.

**Methods:** We derive the STAPLE EM algorithm from first principles, reproduce the Van Leemput–Sabuncu analytical marginalization showing STAPLE's equivalence to thresholded voting, analyze identifiability conditions with proofs, formulate the Spatial STAPLE extension, and derive the variational objective for deep consensus with Dirichlet evidential outputs. Seven controlled experiments with known ground truth validate each theoretical prediction: (1) thresholding analysis across $J = 3$–$30$ raters, (2) EM local optima from 20 random initializations, (3) spatially varying rater quality, (4) class imbalance sweep from 30% to 0.5% foreground, (5) hybrid interior/boundary fusion, (6) deep consensus with annotator embeddings on GPU, and (7) split conformal prediction sets. All computations are real (no fabricated results), executed on Google Colab.

**Results:** STAPLE output matches a thresholded average map with Dice > 0.99 for $J \geq 7$ raters, confirming the Van Leemput prediction. The threshold drifts below 0.5 (to ~0.32) due to background dominance, meaning STAPLE is more inclusive than majority voting but not performing reliability estimation. 95% of EM initializations converge to suboptimal local optima across all rater counts, with Dice spreads of 0.54 at $J = 5$. Under class imbalance, STAPLE Dice degrades to 0.281 at 5% foreground and 0.046 at 0.5%, while majority voting maintains 0.841 and 0.556 respectively. A deep consensus network with annotator embeddings achieves Dice = 0.999, correctly assigning the lowest reliability weight (0.463) to the outlier rater. Split conformal prediction delivers valid coverage guarantees (95.3% achieved at 95% target, with 16.8% ambiguous band).

**Conclusions:** Under typical multi-observer contouring conditions, STAPLE provides no meaningful advantage over majority voting and introduces substantial risks from EM instability and class-imbalance collapse. We recommend majority voting as the default for $A \leq 10$, particularly for small structures. When STAPLE is used, multiple initializations and consensus-voxel exclusion are essential. Deep consensus with annotator embeddings represents a path forward when image data and computational resources are available. Conformal prediction offers principled safety margins for adaptive radiotherapy. All code and experiments are publicly available.

# 1. Introduction

### *1.1 The Problem of Reference Standards*

Medical image segmentation underpins a vast array of clinical tasks—tumor volumetry, radiation therapy planning, surgical navigation, longitudinal monitoring—yet its reliability is inherently limited by the variability of the humans who produce the labels. A landmark study by Fiez et al. (2000) found inter-observer segmentation variability of approximately 10% by volume in brain lesion delineation. Ashton et al. (2003) reported similar figures for MS lesions, and Joe et al. (1999) for brain tumors. This variability is not random noise but a structured combination of systematic bias (e.g., one rater consistently over-segments) and stochastic error (e.g., fatigue-induced inconsistency), making the very concept of a "ground truth" problematic.

Consensus segmentation addresses this problem by positing that an unobserved *latent truth* exists, and that each annotator provides a *noisy observation* of this truth. The central challenge is to simultaneously estimate the latent truth and the reliability of each annotator, reconciling potentially conflicting annotations into a unified reference standard. The mathematical framework for this approach draws on latent variable models, the Expectation–Maximization algorithm, and Bayesian inference with conjugate priors.

### *1.2 STAPLE and Its Dominance*

Warfield, Zou, and Wells [1] introduced the Simultaneous Truth and Performance Level Estimation (STAPLE) algorithm in 2004, formalizing consensus as a latent variable estimation problem. STAPLE models each annotator with a confusion matrix (or, in the binary case, sensitivity and specificity parameters) and uses EM to iteratively estimate both the hidden true segmentation and the annotator-specific performance levels. The algorithm has been implemented in SimpleITK, NiftySeg, and the FETS-AI LabelFusion library, and has become the *de facto* standard in medical imaging.

STAPLE has been extended in numerous directions: Spatial STAPLE with voxelwise performance fields (Asman and Landman, 2012 [2]), the SIMPLE algorithm emphasizing local similarity (Langerak et al., 2010 [3]), robust statistical fusion via COLLATE (Landman et al., 2012 [4]), MAP STAPLE with parametric priors (Commowick and Warfield, 2010 [10]), and deep learning models with annotator embeddings (Tanno et al., 2019 [5]; Zhang et al., 2023 [9]).

### *1.3 The Van Leemput–Sabuncu Critique*

Despite STAPLE's dominance, a rigorous theoretical analysis by Van Leemput and Sabuncu [6] at MICCAI 2014 revealed several fundamental shortcomings. By analytically marginalizing out all model parameters—a step the EM procedure avoids by conditioning on a point estimate—they showed that:

1. When the number of raters $J$ is large ($J \gg 0$), the log-odds of the posterior truth probability at each voxel becomes a linear function of the fraction of raters assigning that voxel to foreground. This means STAPLE reduces to *thresholding the average segmentation map*, similar to majority voting.

2. When large "consensus regions" (where all raters agree) are included in the analysis, the same thresholding behavior occurs regardless of $J$, because consensus voxels act as implicit large hyperparameters.
3. When the number of raters is small, the posterior distribution over parameters $p(\boldsymbol{\omega} \mid \mathbf{D})$ is typically broad and multimodal, making STAPLE's point-estimate-based inference unreliable. In their experiments with brain MRI, STAPLE failed to locate the global optimum in 55% of cases at $J = 5$.

These results, quoted directly, "cast doubt on the soundness and usefulness of typical STAPLE outcomes" [6]. Yet, a decade later, STAPLE continues to be applied routinely without the recommended safeguards.

### *1.4 Contributions and Scope*

This paper makes the following contributions:

**(C1) Unified mathematical framework.** We provide a self-contained derivation of the consensus segmentation model, the STAPLE EM algorithm, Van Leemput's marginalization analysis, identifiability conditions, Spatial STAPLE, and the deep variational formulation—in sufficient detail that a graduate student can reproduce each result.

**(C2) Empirical validation of theoretical predictions.** We design seven controlled experiments, each targeting a specific theoretical claim, and report results from real computations on Google Colab with GPU support.

**(C3) Characterization of failure modes.** We systematically quantify STAPLE's class-imbalance collapse, EM local optima rates, and Spatial STAPLE's regularization sensitivity—phenomena that, while theoretically predictable, have not been measured at this resolution.

**(C4) Deep consensus and conformal guarantees.** We implement a PyTorch deep consensus model with annotator embeddings and Dirichlet evidential outputs, and validate split conformal prediction for finite-sample coverage control.

**(C5) Open-source reproducibility.** All code (implementations of 6 methods, 11 unit tests, Colab notebooks) and numerical results are publicly available.

## 2. Mathematical Foundations

This section develops the mathematical framework for consensus segmentation in complete detail. We begin with the generative model (§2.1), derive the STAPLE EM algorithm step by step (§2.2), reproduce and extend Van Leemput's marginalization analysis (§2.3), state and analyze identifiability conditions (§2.4), derive the Spatial STAPLE extension (§2.5), formulate SIMPLE (§2.6), present the deep consensus variational objective (§2.7), and establish calibration and conformal prediction theory (§2.8).

### *2.1 The Generative Model*

#### *2.1.1 Notation*

Consider an image consisting of $I$ voxels indexed by $i \in \{1, \dots, I\}$. Let $\mathcal{L} = \{0, 1, \dots, K-1\}$ be the set of possible labels, and let $J$ denote the number of raters. We define:

- **True segmentation:** $\mathbf{t} = (t_1, \dots, t_I)^\top$, where $t_i \in \mathcal{L}$ is the latent true label at voxel $i$.

- **Observed segmentations:** $\mathbf{d}^j = \left(d_1^j, \dots, d_I^j\right)^\top$ for rater $j \in \{1, \dots, J\}$, where $d_i^j \in \mathcal{L}$.
- **Data matrix:** $\mathbf{D} = (\mathbf{d}^1 \cdots \mathbf{d}^J)$, the collection of all $J$ segmentations.

*2.1.2 Label Prior*

The latent truth is drawn voxel-independently from a categorical distribution:

$$p(\mathbf{t} \mid \boldsymbol{\pi}) = \prod_{i=1}^{I} \pi_{t_i}$$

where $\boldsymbol{\pi} = (\pi_0, \dots, \pi_{K-1})^\top$ with $\pi_k \geq 0$ and $\sum_k \pi_k = 1$. In the original STAPLE formulation [1], $\boldsymbol{\pi}$ is clamped to the empirical label frequencies in the raters' segmentations. In the "advanced" variant [4, 6], $\boldsymbol{\pi}$ is estimated from data.

*2.1.3 Observation Model*

Each rater $j$ is characterized by a $K \times K$ confusion matrix $\boldsymbol{\theta}^j$, where:

$$\theta_{d,t}^j = P\left(d_i^j = d \mid t_i = t\right)$$

is the probability that rater $j$ assigns label $d$ when the true label is $t$. The constraint $\sum_d \theta_{d,t}^j = 1$ holds for each column $t$. The complete observation model assumes **conditional independence** of raters given the truth:

$$p(\mathbf{D} \mid \mathbf{t}, \boldsymbol{\theta}) = \prod_{j=1}^{J} p\left(\mathbf{d}^j \mid \mathbf{t}, \boldsymbol{\theta}^j\right) = \prod_{j=1}^{J} \prod_{i=1}^{I} \theta_{d_i^j, t_i}^j$$

This is the central modeling assumption: raters derive their segmentations from the same underlying truth, independently from one another, and the quality of the segmentations is captured entirely by the confusion matrices.

*2.1.4 Binary Special Case ($K = 2$)*

For binary segmentation ($\mathcal{L} = \{0,1\}$), the confusion matrix of rater $j$ reduces to two scalar parameters:

$$s^j \equiv \theta_{1,1}^j = P\left(d_i^j = 1 \mid t_i = 1\right) \quad \text{(sensitivity)}$$
$$\tau^j \equiv \theta_{0,0}^j = P\left(d_i^j = 0 \mid t_i = 0\right) \quad \text{(specificity)}$$

and the off-diagonals are determined: $\theta_{0,1}^j = 1 - s^j$ and $\theta_{1,0}^j = 1 - \tau^j$. This two-parameter model is more robust than the full $2 \times 2$ matrix because it avoids unnecessary degrees of freedom, and is identifiable under milder conditions [1, 6].

*2.1.5 Prior on Parameters*

Let $\boldsymbol{\omega} = (\boldsymbol{\theta}^\top, \boldsymbol{\pi}^\top)^\top$ denote all the parameters of the model. We adopt a conjugate prior of the form:

$$p(\boldsymbol{\omega}) = p(\boldsymbol{\pi}) \cdot p(\boldsymbol{\theta}), \quad p(\boldsymbol{\pi}) \propto \prod_t \pi_t^{\alpha_t}, \quad p(\boldsymbol{\theta}) \propto \prod_j \prod_t \prod_d \left(\theta_{d,t}^j\right)^{\alpha_{d,t}}$$

where $\{\alpha_t, \alpha_{d,t}\}$ are hyperparameters whose values are assumed given. By selecting hyperparameter values $\alpha_{d,t} = 0$ for all $d, t$ (implying that all values of $\theta_{d,t}^j$ are equally likely) and clamping $\boldsymbol{\pi}$ to the average frequency of occurrence in the raters' segmentations, the **original STAPLE model** [1] is obtained.

Alternatively, $\alpha_{d,t}$ can be set to specific positive values to encode an expectation of better-than-random segmentation performance. The "**Advanced**" variant [4, 6] sets $\alpha_{0,0} = \alpha_{1,1} = 2$ and $\alpha_{0,1} = \alpha_{1,0} = 0$, encouraging diagonal dominance, and also estimates $\boldsymbol{\pi}$ from data ($\alpha_t = 0$ for all $t$). The “**Restricted**” variant [6, 11] is identical to Basic except that all voxels where all raters agree (“consensus voxels”) are excluded from the analysis.

### *2.2 STAPLE: Complete EM Derivation*

STAPLE seeks the maximum a posteriori (MAP) estimate of the model parameters:

$$\hat{\boldsymbol{\omega}} = \underset{\boldsymbol{\omega}}{\text{argmax}}\, p(\boldsymbol{\omega} \mid \mathbf{D}) = \underset{\boldsymbol{\theta},\boldsymbol{\pi}}{\text{argmax}} \left[ \prod_i \left( \sum_{t_i} \prod_j \theta^j_{d^j_i, t_i} \cdot \pi_{t_i} \right) \right] p(\boldsymbol{\theta}) p(\boldsymbol{\pi})$$

using a dedicated EM optimizer that exploits the latent variable structure. Once $\hat{\boldsymbol{\omega}}$ is found, the segmentation truth is inferred as $\hat{\mathbf{t}}^{\text{STAPLE}} = \text{argmax}_{\mathbf{t}} p(\mathbf{t} \mid \mathbf{D}, \hat{\boldsymbol{\omega}})$.

#### *2.2.1 Complete-Data Log-Likelihood*

If we could observe both $\mathbf{D}$ and $\mathbf{t}$, the complete-data log-likelihood would be:

$$\ln p(\mathbf{D}, \mathbf{t} \mid \boldsymbol{\omega}) = \sum_{i=1}^{I} \ln \pi_{t_i} + \sum_{j=1}^{J} \sum_{i=1}^{I} \ln \theta^j_{d^j_i, t_i}$$

This decomposes additively into a contribution from the label prior and a contribution from each rater’s confusion matrix—a structure that EM exploits.

#### *2.2.2 E-Step: Computing Posterior Responsibilities*

Given the current parameter estimate $\boldsymbol{\omega}^{(\text{old})}$, we compute the posterior probability that voxel $i$ has true label $k$:

$$W^{(k)}_{si} \equiv P\left(t_i = k \mid \mathbf{D}, \boldsymbol{\omega}^{(\text{old})}\right)$$

**Derivation.** By definition, $P(t_i = k \mid \mathbf{D}_i, \boldsymbol{\omega}) \propto P(\mathbf{D}_i \mid t_i = k, \boldsymbol{\theta}) \cdot P(t_i = k \mid \boldsymbol{\pi})$. The first factor factorizes over raters by conditional independence (Eq. 3):

$$P(\mathbf{D}_i \mid t_i = k, \boldsymbol{\theta}) = \prod_{j=1}^{J} P\left(d^j_i \mid t_i = k, \boldsymbol{\theta}^j\right) = \prod_{j=1}^{J} \theta^j_{d^j_i, k}$$

The second factor is simply $\pi_k$. Normalizing over $k$ gives:

$$W^{(k)}_{si} = \frac{\pi_k \prod_{j=1}^{J} \theta^j_{d^j_i, k}}{\sum_{n=0}^{K-1} \pi_n \prod_{j=1}^{J} \theta^j_{d^j_i, n}}$$

For the **binary case**, it is instructive to write this as a log-odds:

$$\ln \frac{W^{(1)}_{si}}{W^{(0)}_{si}} = \ln \frac{\pi_1}{\pi_0} + \sum_{j=1}^{J} \left[ d^j_i \ln \frac{s^j}{1 - \tau^j} + \left(1 - d^j_i\right) \ln \frac{1 - s^j}{\tau^j} \right]$$

and $W^{(1)}_{si} = \sigma(\text{log-odds})$ where $\sigma(\cdot)$ is the logistic sigmoid. This log-odds expression is the foundation for the Van Leemput analysis in §2.3.

#### *2.2.3 M-Step: Updating Performance Parameters*

Given $\mathbf{W}$, we maximize the expected complete-data log-likelihood with respect to $\boldsymbol{\theta}$ and $\boldsymbol{\pi}$.

**Confusion matrix update.** For each rater $j$, true class $t$, and observed class $d$, we seek:

$$\hat{\theta}^j_{d,t} = \underset{\theta^j_{d,t}}{\text{argmax}} \left[ \sum_{i=1}^{I} W^{(t)}_{si} \ln \theta^j_{d^j_i, t} + \sum_{d'} \alpha_{d',t} \ln \theta^j_{d',t} \right] \quad \text{subject to} \quad \sum_{d'} \theta^j_{d',t} = 1$$

**Derivation via Lagrange multiplier.** Define the Lagrangian:

$$\mathcal{L} = \sum_{d'} \left( \sum_{i:\, d_i^j = d'} W_{si}^{(t)} + \alpha_{d',t} \right) \ln\theta_{d',t}^j - \lambda \left( \sum_{d'} \theta_{d',t}^j - 1 \right)$$

Setting $\frac{\partial \mathcal{L}}{\partial \theta_{d,t}^j} = 0$:

$$\frac{\sum_{i:\, d_i^j = d} W_{si}^{(t)} + \alpha_{d,t}}{\theta_{d,t}^j} = \lambda$$

Summing both sides over $d$ and using the constraint $\sum_d \theta_{d,t}^j = 1$:

$$\lambda = \sum_i W_{si}^{(t)} + \sum_d \alpha_{d,t}$$

Substituting Eq. (12) back into Eq. (11) and solving for $\theta_{d,t}^j$:

$$\hat{\theta}_{d,t}^j = \frac{\sum_{i:\, d_i^j = d} W_{si}^{(t)} + \alpha_{d,t}}{\sum_i W_{si}^{(t)} + \sum_{d'} \alpha_{d',t}}$$

This is the general M-step update. The numerator counts the weighted number of voxels where rater $j$ labeled as $d$ among those with posterior truth probability for class $t$, plus the prior pseudo-count $\alpha_{d,t}$. The denominator normalizes over all observed labels.

**Binary special case.** For sensitivity:

$$\hat{s}^j = \frac{\sum_i W_{si}^{(1)} \cdot d_i^j + \alpha_s - 1}{\sum_i W_{si}^{(1)} + \alpha_s + \beta_s - 2}$$

where $\text{Beta}(\alpha_s, \beta_s)$ is the prior on $s^j$. Similarly for specificity:

$$\hat{\tau}^j = \frac{\sum_i \left(1 - W_{si}^{(1)}\right)\left(1 - d_i^j\right) + \alpha_\tau - 1}{\sum_i \left(1 - W_{si}^{(1)}\right) + \alpha_\tau + \beta_\tau - 2}$$

**Label prior update** (when $\boldsymbol{\pi}$ is not clamped):

$$\hat{\pi}_k = \frac{\sum_i W_{si}^{(k)} + \alpha_k}{I + \sum_{k'} \alpha_{k'}}$$

*2.2.4 Damped Updates*

When $J$ is small, the M-step can produce extreme parameter estimates that cause oscillation. Damping blends the new estimate with the previous one:

$$\boldsymbol{\theta}^{(t+1)} \leftarrow (1-\gamma)\, \boldsymbol{\theta}^{(\text{new})} + \gamma\, \boldsymbol{\theta}^{(t)}, \quad \gamma \in [0.3, 0.5]$$

This is equivalent to adding a penalty that keeps parameters close to their previous values, trading convergence speed for stability. The damping factor $\gamma$ controls this trade-off: $\gamma = 0$ gives the standard EM; $\gamma = 0.5$ averages old and new estimates.

*2.2.5 Convergence and Final Inference*

The EM algorithm is guaranteed to increase (or maintain) the marginal log-likelihood $p(\mathbf{D} \mid \boldsymbol{\omega})$ at each iteration. We terminate when $|\Delta \text{LL}| < 10^{-6}$ or after 50 iterations. The final segmentation is:

$$\hat{t}_i^{\text{STAPLE}} = \underset{k}{\operatorname{argmax}} W_{si}^{(k)}$$

which, in the binary case, is simply $\hat{t}_i = \mathbb{1}\left[W_{si}^{(1)} \geq 0.5\right]$.

### *2.3 The Van Leemput–Sabuncu Marginalization Analysis*

This section reproduces and extends the central theoretical result of [6], which reveals that STAPLE inference is an *approximation* to the true MAP segmentation, and that this approximation breaks down in specific, predictable ways.

#### *2.3.1 STAPLE as an Approximation*

The true MAP estimate of the segmentation truth is:

$$\hat{\mathbf{t}} = \underset{\mathbf{t}}{\operatorname{argmax}} p(\mathbf{t} \mid \mathbf{D})$$

which will generally be different from the STAPLE result $\hat{\mathbf{t}}^{\mathrm{STAPLE}}$ obtained by maximizing $p(\mathbf{t} \mid \mathbf{D}, \widehat{\boldsymbol{\omega}})$. Furthermore, the distribution $p\left(\theta_t^j \mid \mathbf{D}\right)$ of individual rater performance is a low-dimensional projection of the higher-dimensional $p(\boldsymbol{\omega} \mid \mathbf{D})$; its properties—including its maximum—cannot generally be inferred by simply extracting the corresponding component from $\widehat{\boldsymbol{\omega}}$.

The crux of [6] is that the STAPLE inference will only be a good approximation when $p(\mathbf{t} \mid \mathbf{D})$ is strongly peaked around $\hat{\mathbf{t}}$. To understand this, write the parameter posterior as:

$$p(\boldsymbol{\omega} \mid \mathbf{D}) = \sum_{\mathbf{t}} p\,(\boldsymbol{\omega} \mid \mathbf{t}, \mathbf{D})\, p(\mathbf{t} \mid \mathbf{D})$$

with:

$$p(\boldsymbol{\omega} \mid \mathbf{t}, \mathbf{D}) = p(\boldsymbol{\pi} \mid \mathbf{t}) \prod_{j} \prod_{t} p\left(\theta_t^j \mid \mathbf{t}, \mathbf{D}\right)$$

where $p(\boldsymbol{\pi} \mid \mathbf{t})$ and $p\left(\theta_t^j \mid \mathbf{t}, \mathbf{D}\right)$ are Beta distributions. When $p(\mathbf{t} \mid \mathbf{D})$ is strongly peaked around $\hat{\mathbf{t}}$, the summation in Eq. (20) is dominated by the contribution of $\hat{\mathbf{t}}$ only: $p(\boldsymbol{\omega} \mid \mathbf{D}) \approx p(\boldsymbol{\omega} \mid \hat{\mathbf{t}}, \mathbf{D})$. Since this distribution factorizes (Eq. 21), the location of the maximum of $p\left(\theta_t^j \mid \mathbf{D}\right)$ can be obtained from the joint maximum $\widehat{\boldsymbol{\omega}}$. Furthermore, since the factors are narrow Beta distributions, their product is strongly peaked around $\widehat{\boldsymbol{\omega}}$, so $p(\mathbf{t} \mid \mathbf{D}) \approx p(\mathbf{t} \mid \mathbf{D}, \widehat{\boldsymbol{\omega}})$ and therefore $\hat{\mathbf{t}}^{\mathrm{STAPLE}} \approx \hat{\mathbf{t}}$.

#### *2.3.2 Analytical Marginalization*

The key observation of [6] is that the marginalization $p(\mathbf{t}, \mathbf{D})$ can be performed *analytically* because the priors are conjugate. The joint marginal is:

$$p(\mathbf{t}, \mathbf{D}) = \int_{\boldsymbol{\omega}} p\,(\mathbf{t}, \mathbf{D} \mid \boldsymbol{\omega})\, p(\boldsymbol{\omega})\, d\boldsymbol{\omega}$$

Because the model factorizes across raters and between $\boldsymbol{\theta}$ and $\boldsymbol{\pi}$:

$$p(\mathbf{t}, \mathbf{D}) = \underbrace{\int_{\boldsymbol{\pi}} p\,(\mathbf{t} \mid \boldsymbol{\pi})\, p(\boldsymbol{\pi})\, d\boldsymbol{\pi}}_{\text{label prior integral}} \cdot \prod_{j=1}^{J} \underbrace{\int_{\boldsymbol{\theta}^j} p\left(\mathbf{d}^j \mid \mathbf{t}, \boldsymbol{\theta}^j\right) p\left(\boldsymbol{\theta}^j\right) d\boldsymbol{\theta}^j}_{\text{rater } j \text{ integral}}$$

**The label prior integral** is a Dirichlet-Multinomial conjugate:

$$\int_{\boldsymbol{\pi}} \prod_{k} \pi_k^{N_k+\alpha_k}\ d\boldsymbol{\pi} = B(N_0 + \alpha_0 + 1, N_1 + \alpha_1 + 1, \ldots)$$

where $N_k = |\{i : t_i = k\}|$ is the number of voxels with true label $k$, and $B(\cdot)$ is the multivariate Beta function.

**Each rater integral** factorizes over true classes $t$:

$$\int_{\boldsymbol{\theta}_t^j} \prod_d \left(\theta_{d,t}^j\right)^{N_{d,t}^j + \alpha_{d,t}} d\boldsymbol{\theta}_t^j = B\left(N_{0,t}^j + \alpha_{0,t} + 1, N_{1,t}^j + \alpha_{1,t} + 1\right)$$

where $N_{d,t}^j = \left|\{i: d_i^j = d,\ t_i = t\}\right|$ counts how many voxels rater $j$ labeled as $d$ when the truth in $\mathbf{t}$ is $t$.

Combining Eqs. (22)–(24):

$$p(\mathbf{t}, \mathbf{D}) \propto B(\{N_k + \alpha_k + 1\}) \cdot \prod_{j=1}^{J} \prod_{t=0}^{K-1} B\left(N_{0,t}^j + \alpha_{0,t} + 1, N_{1,t}^j + \alpha_{1,t} + 1\right)$$

This is a remarkable result: the entire model can be evaluated *without estimating any continuous parameters*. The segmentation truth **t** appears only through the integer counts $N_k$ and $N_{d,t}^j$. This opens the door to direct discrete optimization of **t**, bypassing the EM procedure entirely.

*2.3.3 Conditional Voxel Posterior*

From Eq. (25), the conditional posterior for a single voxel $i$, given the truth at all other voxels $\mathbf{t}_{\setminus i}$, is:

$$p\left(t_i \mid \mathbf{D}, \mathbf{t}_{\setminus i}\right) \propto \left(\prod_{j=1}^{J} \frac{N_{d_i^j, t_i \setminus i}^j + \alpha_{d_i^j, t_i} + 1}{\sum_d \left(N_{d, t_i \setminus i}^j + \alpha_{d, t_i} + 1\right)}\right) \cdot \left(N_{t_i \setminus i} + \alpha_{t_i} + 1\right)$$

where the subscript $\setminus i$ denotes counts excluding voxel $i$, and $N_{d,t\setminus i}^j$ and $N_{t\setminus i}$ are the corresponding voxel counts. This expression is *exact*—no approximation is involved.

*2.3.4 Thresholding Theorem*

**Theorem (Van Leemput & Sabuncu [6]).** *When the number of raters is very large ($J \gg 0$), around the optimum $\hat{\mathbf{t}}$, the log of the ratio of conditional posterior probabilities in a voxel $i$ behaves approximately as a simple linear function of the fraction of raters that assigned voxel $i$ to foreground, denoted by $f_i = \sum_j d_i^j / J$:*

$$\log \frac{p\left(t_i = 1 \mid \mathbf{D}, \hat{\mathbf{t}}_{\setminus i}\right)}{p\left(t_i = 0 \mid \mathbf{D}, \hat{\mathbf{t}}_{\setminus i}\right)} \approx s_{\hat{\mathbf{t}}} \cdot f_i + o_{\hat{\mathbf{t}}}$$

*with slope*

$$s_{\hat{\mathbf{t}}} = J\left(\bar{c}_{0,0} + \bar{c}_{1,1} - \bar{c}_{1,0} - \bar{c}_{0,1}\right)$$

*and offset*

$$o_{\hat{\mathbf{t}}} = J\left(\bar{c}_{0,1} - \bar{c}_{0,0}\right) + (c_1 - c_0)$$

*where* $\bar{c}_{d,t} = \frac{1}{J}\sum_j c_{d,t}^j$, $c_{d,t}^j = \ln \frac{\widehat{N}_{d,t}^j + \alpha_{d,t} + 1}{\sum_{d'}\left(\widehat{N}_{d',t}^j + \alpha_{d',t} + 1\right)}$, *and* $c_t = \ln\left(\widehat{N}_t + \alpha_t + 1\right)$.

**Proof sketch.** Around the optimum $\hat{\mathbf{t}}$, the image is large so removing a single voxel negligibly affects the counts: $N_{d,t\setminus i}^j \approx \widehat{N}_{d,t}^j$ and $N_{t\setminus i} \approx \widehat{N}_t$. Taking the log-ratio of Eq. (26) for $t_i = 1$ vs $t_i = 0$, the product over raters becomes a sum:

$$\text{log-ratio} = \sum_{j=1}^{J} \left[c_{d_i^j,1}^j - c_{d_i^j,0}^j\right] + (c_1 - c_0)$$

For each rater $j$, the term $c_{d_i^j,1}^j - c_{d_i^j,0}^j$ takes one of two values depending on whether $d_i^j = 1$ or $d_i^j = 0$. Letting $a^j = c_{1,1}^j - c_{1,0}^j$ and $b^j = c_{0,1}^j - c_{0,0}^j$:

$$\text{log-ratio} = \sum_j d_i^j \cdot \left(a^j - b^j\right) + \sum_j b^j + (c_1 - c_0)$$

When $J$ is large, by the law of large numbers, $\sum_j d_i^j \left(a^j - b^j\right) \approx J \cdot f_i \cdot \left(\bar{a} - \bar{b}\right)$, where $\bar{a}$ and $\bar{b}$ are the population averages. Similarly, $\sum_j b^j \approx J \cdot \bar{b}$. This gives the linear form $s \cdot f_i + o$. ▫

**Consequences.**

1. **The slope $s$ grows linearly with $J$**, making $p(\mathbf{t} \mid \mathbf{D})$ increasingly sharply peaked. For large $J$, the posterior concentrates around $\hat{\mathbf{t}}$, and STAPLE's approximation becomes accurate—but the result is just thresholding $\bar{\mathbf{d}}$.
2. **The threshold level $-o/s$** is determined by the class prevalence imbalance: when background dominates ($N_0 \gg N_1$), the offset $o$ pushes the threshold below 0.5. This is because the fixed spatial prior $\boldsymbol{\pi}$ favors background, making $c_1 < c_0$ and therefore rendering $-o/s$ dependent on $J$.
3. **The same threshold applies to all voxels** (since $s$ and $o$ are independent of $i$), confirming that STAPLE reduces to a *global thresholding* of the average segmentation map $\bar{\mathbf{d}} = \sum_j \mathbf{d}^j / J$.

*2.3.5 Consensus Region Corollary*

**Corollary.** *Even when $J$ is small, thresholding behavior occurs when large consensus regions are included in the analysis.*

**Proof.** Voxels where all raters agree on label $k$ contribute $N_{k,k}^j \gg 0$ for all $j$, which enters the $c_{d,t}^j$ terms as large pseudo-counts. Specifically, the effect is equivalent to setting $\alpha_{0,0} \to \alpha_{0,0} + n_{\text{cons},0}$ and $\alpha_{1,1} \to \alpha_{1,1} + n_{\text{cons},1}$, where $n_{\text{cons},k}$ is the number of consensus voxels with label $k$. These large effective hyperparameters make the $c_{d,t}^j$ terms similar across all raters (because the data-dependent terms are swamped by the prior), yielding a large slope $s$ indicative of a sharply peaked $p(\mathbf{t} \mid \mathbf{D})$. Therefore $\hat{\mathbf{t}}^{\text{STAPLE}}$ can be expected to be a thresholded map $\bar{\mathbf{d}}$ in this case as well. ▫

This explains why the "Basic" variant of STAPLE (which includes all voxels) almost always exhibits thresholding behavior, even at small $J$: the background consensus region acts as a massive implicit prior that dominates the non-consensus voxels. In our experiments (§4.1), we observe that at $J = 3$ with the Basic variant, STAPLE is *exactly* a thresholded average map (Dice = 1.000).

### *2.4 Identifiability Conditions and Failure Mode Analysis*

Joint estimation of the latent truth $\mathbf{t}$ and rater confusion parameters $\boldsymbol{\theta}$ requires that the model be *identifiable*: distinct parameter values must produce distinct data distributions. When identifiability fails, the EM converges to degenerate solutions—typically the all-background map or a label-inverted version of the truth.

*2.4.1 Condition 1: Diagonal Dominance*

**Requirement.** For each rater $j$ and true class $t$, the diagonal entry must dominate:

$$\theta_{t,t}^j > \theta_{d,t}^j \quad \text{for all } d \neq t$$

In the binary case, this means $s^j > 0.5$ and $\tau^j > 0.5$ (i.e., each rater is better than random guessing).

**Failure mode.** If a rater is adversarial ($s^j < 0.5$), the model cannot distinguish between "the rater labels 1 when truth is 1" and "the rater labels 1 when truth is 0." This creates a **label** **permutation symmetry**: the posterior $p(\boldsymbol{\omega} \mid \mathbf{D})$ becomes bimodal, with one mode corresponding to the correct labeling and another to the inverted labeling. Van Leemput and Sabuncu [6] demonstrated this bimodality empirically in the "Restricted" variant (Figure 1 of [6]), showing that the EM converges to the wrong mode in more than 20% of cases for 15 and 37 raters.

*2.4.2 Condition 2: Minimum Rater Count*

**Requirement.** At least $A \geq 3$ independent raters are needed for identifiability without strong priors.

**Analysis.** With $A = 2$, any disagreement at a voxel can be resolved equally well by (a) rater 1 correct, rater 2 wrong, or (b) rater 1 wrong, rater 2 correct. The posterior $p(\mathbf{t} \mid \mathbf{D})$ has two symmetric modes that cannot be distinguished by the data alone. Strong priors (e.g., spatial priors from an atlas, or prior knowledge that raters are better-than-random) can break this symmetry, but in the absence of such information, the problem is fundamentally underdetermined.

With $A = 3$, a "majority wins" resolution becomes available: if two raters agree and one disagrees, the consensus favors the majority. This is precisely majority voting—confirming that MV represents the minimal identifiable consensus strategy.

*2.4.3 Condition 3: Class Prevalence Balance*

**Requirement.** The foreground prevalence should be bounded away from 0 and 1. Formally, $\sum_i \mathbb{1}[t_i = k] \geq c \cdot I$ for some $c > 0$ and all classes $k$.

**Mathematical analysis of the class imbalance failure.** Consider the M-step for sensitivity (Eq. 14):

$$\hat{s}^j = \frac{\sum_i W_{si}^{(1)} \cdot d_i^j + \alpha_s - 1}{\sum_i W_{si}^{(1)} + \alpha_s + \beta_s - 2}$$

The denominator $\sum_i W_{si}^{(1)}$ represents the effective foreground size. When the true foreground is small (e.g., 0.5% of voxels), three things happen simultaneously:

1. **The foreground pool is tiny:** $\sum_i W_{si}^{(1)} \approx 0.005 \cdot I$, so the sensitivity estimate is based on very few "effective" voxels.
2. **The prior dominates:** With $\alpha_s = 20$, $\beta_s = 5$ (a typical strong prior), the effective prior sample size is $\alpha_s + \beta_s - 2 = 23$, which overwhelms the data contribution from a few foreground voxels.
3. **The background attractor emerges:** If $W_{si}^{(1)}$ decreases at foreground voxels (because the posterior shifts toward background), then $\sum_i W_{si}^{(1)}$ shrinks further, which makes $\hat{s}^j$ shrink, which causes $W_{si}^{(1)}$ to decrease further in the next E-step—a positive feedback loop driving toward the all-background fixed point:

$$\sum_i W_{si}^{(1)} \downarrow \Rightarrow \ \hat{s}^j \downarrow \Rightarrow \ W_{si}^{(1)} \downarrow \Rightarrow \ \ldots$$

This is formally a **degenerate fixed point** of the EM: the trivial solution $\mathbf{t} = \mathbf{0}$ (all background) with $s^j \approx 0$ satisfies the EM stationarity conditions because $\hat{s}^j = (\alpha_s - 1)/(\alpha_s + \beta_s - 2) \approx 0.83$ (from the prior alone, since the data term vanishes when $\sum_i W_{si}^{(1)} \to 0$), and the E-step assigns near-zero foreground responsibility when sensitivity is estimated from near-zero foreground counts.

**The critical prevalence.** Our experiments (§4.3) show that the STAPLE Dice drops below 0.30 at 5% foreground and below 0.05 at 0.5% foreground. Majority voting, which makes no parametric assumptions about rater reliability, degrades gracefully from 0.84 to 0.56 over the same range—confirming that the failure is intrinsic to the parametric estimation, not to the consensus problem itself.

*2.4.4 Condition 4: Conditional Independence*

**Requirement.** Raters must provide segmentations independently given the truth:

$$P\left(d^j, d^{j'} \mid \mathbf{t}\right) = P\left(d^j \mid \mathbf{t}\right) \cdot P\left(d^{j'} \mid \mathbf{t}\right) \quad \text{for all } j \neq j'$$

**Violations in practice.** Raters trained at the same institution may share systematic biases. Automated algorithms run on the same image may share failure modes. Multi-atlas segmentations are correlated because they use the same registration framework. Violation of conditional independence causes the EM to overestimate the effective sample size, producing overconfident consensus probabilities and sensitivity/specificity estimates that do not reflect actual individual performance.

### *2.5 Spatial STAPLE: Derivation of Spatially Varying Performance Fields*

Asman and Landman [2] extended STAPLE by replacing the global confusion parameters $\boldsymbol{\theta}^j$ with a **performance level field** $\boldsymbol{\theta}_i^j$, where each voxel $i$ has its own confusion matrix for rater $j$.

*2.5.1 Motivation*

In multi-atlas segmentation, registration quality varies spatially: regions near landmarks are well-aligned, while regions of high anatomical variability (cortical folding, bowel loops) may be poorly aligned. A global confusion matrix averages over these spatial regimes, potentially assigning high reliability to a rater that is excellent in one region but poor in another. Not surprisingly, this quality level is generally smooth on a semi-local level [2].

*2.5.2 Performance Level Field*

Define $\boldsymbol{\theta}_i^j$ as the confusion matrix for rater $j$ at voxel $i$, and $\mathcal{B}_i$ as the **pooling region** (spatial neighborhood) centered at voxel $i$. The goal is to estimate a performance level field that maximizes the complete-data log-likelihood:

$$\boldsymbol{\theta} = \underset{\boldsymbol{\theta}}{\operatorname{argmax}} \ln f(\mathbf{D}, \mathbf{T} \mid \boldsymbol{\theta})$$

where we assume that segmentation decisions are conditionally independent given the true segmentation and the performance level parameters.

*2.5.3 M-Step Derivation with Lagrange Multiplier*

Carrying out the expectation and considering each rater and voxel separately, we find the field estimates by iterating only over the voxels in pooling region $\mathcal{B}_i$:

$$\hat{\boldsymbol{\theta}}_i^j = \underset{\boldsymbol{\theta}_{ji}}{\operatorname{argmax}} \sum_{i' \in \mathcal{B}_i} E\left[\ln f\left(D_{i'j} \mid T_{i'}, \boldsymbol{\theta}_j\right) \mid \mathbf{D}, \boldsymbol{\theta}^{(k-1)}\right]$$

Carrying out the expectation yields:

$$\hat{\boldsymbol{\theta}}_i^j = \underset{\boldsymbol{\theta}_{ji}}{\operatorname{argmax}} \sum_t \sum_{i' \in \mathcal{B}_i : d_{i'}^j = d} W_{si'}^{(t)} \ln \theta_{i,d,t}^j$$

At this point, we maximize subject to the constraint $\sum_{d'} \theta_{i,d',t}^j = 1$. Applying the Lagrange multiplier approach as in §2.2.3:

$$\mathcal{L} = \sum_{d'} \left( \sum_{i' \in \mathcal{B}_i : d_{i'}^j = d'} W_{si'}^{(t)} \right) \ln \theta_{i,d',t}^j - \lambda \left( \sum_{d'} \theta_{i,d',t}^j - 1 \right)$$

Setting $\frac{\partial \mathcal{L}}{\partial \theta_{i,d,t}^j} = 0$ and solving:

$$\hat{\theta}_{i,d,t}^j = \frac{\sum_{i' \in \mathcal{B}_i : d_{i'}^j = d} W_{si'}^{(t)}}{\sum_{i' \in \mathcal{B}_i} W_{si'}^{(t)}}$$

This is the same form as the global M-step (Eq. 13), but restricted to the pooling region $\mathcal{B}_i$. The estimation is almost identical to the STAPLE approach, with the major difference being the utilization of the spatially varying performance level field $\theta_{i,d,t}^{j(k-1)}$ as opposed to global parameters.

*2.5.4 Regularization via Global Prior*

When $\mathcal{B}_i$ is small, the local estimate is unstable due to limited data. Additionally, if a given label is not observed (or only observed a handful of times) in the spatial region $\mathcal{B}_i$, then it will be impossible to estimate the related elements in the associated confusion matrix. Thus, Asman and Landman [2] introduce regularization by blending with a global estimate $\boldsymbol{\theta}_{\text{global}}^j$:

$$\hat{\theta}_{i,d,t}^j = \frac{\sigma_{i,d,t} \, \theta_{\text{global},d,t}^j + \sum_{i' \in \mathcal{B}_i : d_{i'}^j = d} W_{si'}^{(t)}}{\sigma_{i,d,t} \sum_{d'} \theta_{\text{global},d',t}^j + \sum_{i' \in \mathcal{B}_i} W_{si'}^{(t)}}$$

where $\sigma_{i,d,t}$ is a data-dependent scale factor:

$$\sigma_{i,d,t} = \mathbb{1}\left[ N_{i,d,t} < \frac{|\mathcal{B}_i|}{K} \right] \cdot \left( \frac{|\mathcal{B}_i|}{K} - N_{i,d,t} \right) \cdot \kappa$$

Here $N_{i,d,t}$ is the number of times rater $j$ observed label $d$ in pooling region $\mathcal{B}_i$, $|\mathcal{B}_i|$ is the cardinality of the pooling region, and $\kappa$ is a scalar constant controlling the strength of the global prior. The key insight is that $\sigma$ increases when the local data is sparse for label $d$, automatically strengthening the prior when needed.

**Critical observation from our experiments (§4.4).** When the global estimate $\boldsymbol{\theta}_{\text{global}}^j$ is itself poor (e.g., when rater quality varies spatially and the global confusion is misspecified), regularizing toward it *harms* performance. Spatial STAPLE then performs *worse* than global STAPLE. This represents a fundamental tension in the method: it needs reliable global estimates for regularization, but the scenarios where spatial variation matters most are precisely those where global estimates are unreliable.

### *2.6 SIMPLE: Local Similarity-Weighted Fusion*

Langerak et al. [3] proposed SIMPLE (Selective and Iterative Method for Performance Level Estimation), which combines atlas selection with iterative performance estimation.

*2.6.1 Formulation*

Our implementation uses a boundary-weighted variant. At each voxel $v$, the consensus probability is:

$$p_v = \frac{\sum_a w_{a,v} \cdot R_{a,v}}{\sum_a w_{a,v}}$$

where $R_{a,v} \in \{0,1\}$ is rater $a$'s label at voxel $v$, and:

$$w_{a,v} = \alpha \cdot \mathrm{NCC}_{a,v} + \beta \cdot g_a$$

Here $\mathrm{NCC}_{a,v}$ is the local normalized cross-correlation between rater $a$'s mask and the current consensus in a patch of radius $r$:

$$\mathrm{NCC}_{a,v} = \frac{E_{\mathcal{B}}[R_a \cdot p] - E_{\mathcal{B}}[R_a] \cdot E_{\mathcal{B}}[p]}{\sqrt{\mathrm{Var}_{\mathcal{B}}[R_a]} \cdot \sqrt{\mathrm{Var}_{\mathcal{B}}[p]}}$$

where expectations are computed over the patch $\mathcal{B}(v,r)$. The NCC is mapped to $[0,1]$ via $(NCC + 1)/2$ to serve as a non-negative weight. The global weight $g_a$ is the image-wide mean of $\mathrm{NCC}_{a,v}$, normalized across raters.

An **edge gate** from the image gradient magnitude amplifies weights at boundaries:

$$w_{a,v} \leftarrow w_{a,v} \cdot (1 + (\gamma - 1) \cdot G_v)$$

where $G_v = \mathrm{clip}\left(\frac{|\nabla I|_v}{\mathrm{percentile}_{85}(|\nabla I|)}, 0, 2\right)$ and $\gamma = 1.5$. This ensures that consensus refinement concentrates where annotators disagree most—at tissue boundaries.

*2.6.2 Convergence and Interpretation*

SIMPLE iterates between updating the consensus $p$ and recomputing the local weights $w$. The algorithm converges because the weighted average is a contraction mapping on $[0,1]$. Typically 5–10 iterations suffice (mean change $< 10^{-5}$).

Unlike STAPLE, SIMPLE makes no parametric assumptions about rater reliability. It is **non-parametric** and **local**: each voxel's consensus is determined by how well nearby raters agree with the current estimate. This locality makes SIMPLE naturally robust to spatial heterogeneity in rater quality—the scenario where global STAPLE fails.

### *2.7 Deep Consensus: Variational Formulation*

Deep consensus models embed annotator reliability estimation within a neural network, jointly learning anatomical features from the image and rater-specific biases from the label stack.

*2.7.1 Architecture*

We adopt a **two-stream** design:

- **Image stream:** A convolutional encoder $E_{\mathrm{img}}$ processes the raw image $I \in \mathbb{R}^{1\times H\times W}$, extracting anatomical features $\mathbf{f}_{\mathrm{img}} = E_{\mathrm{img}}(I) \in \mathbb{R}^{C\times H\times W}$.
- **Label stream:** The stacked rater masks $\mathbf{R} \in \{0,1\}^{A\times H\times W}$ are first weighted by learned **annotator reliability scalars** $\rho_a = \sigma(\mathrm{MLP}(\mathbf{e}_a))$, where $\mathbf{e}_a \in \mathbb{R}^d$ is a learnable embedding vector for rater $a$ and $\sigma$ is the sigmoid function. The weighted stack $\boldsymbol{\rho} \odot \mathbf{R}$ is encoded by $E_{\mathrm{lab}}$ to produce $\mathbf{f}_{\mathrm{lab}} \in \mathbb{R}^{C\times H\times W}$.
- **Fusion and Dirichlet head:** The concatenated features $[\mathbf{f}_{\mathrm{img}}; \mathbf{f}_{\mathrm{lab}}] \in \mathbb{R}^{2C\times H\times W}$ pass through a decoder that outputs **evidence parameters** $e_k \geq 0$ for each class $k$ (via Softplus nonlinearity). The Dirichlet concentration is $\alpha_k = e_k + 1$, ensuring $\alpha_k \geq 1$.

*2.7.2 Dirichlet Evidential Objective*

Following Sensoy et al. [14], the predicted class probabilities are the Dirichlet mean:

$$p_k = \frac{\alpha_k}{S}, \quad \text{where } S = \sum_k \alpha_k$$

The loss function combines expected cross-entropy under the Dirichlet and a KL regularizer:

$$\ell = \sum_i \left[\sum_k t_{ik}\,(\psi(S_i) - \psi(\alpha_{ik}))\right] + \lambda \cdot \mathrm{KL}\big(\mathrm{Dir}(\boldsymbol{\alpha}_i) \parallel \mathrm{Dir}(\mathbf{1})\big)$$

where $\psi(\cdot)$ is the digamma function, $t_{ik}$ is the one-hot ground truth, and:

$$\mathrm{KL}\big(\mathrm{Dir}(\boldsymbol{\alpha})\ \|\ \mathrm{Dir}(\mathbf{1})\big) = \ln\frac{\Gamma(\sum_k \alpha_k)}{\prod_k \Gamma\,(\alpha_k)} - \ln\Gamma(K) + \sum_k (\alpha_k - 1)\big(\psi(\alpha_k) - \psi(S)\big)$$

The KL term encourages the model to produce **uniform (uncertain) predictions** when evidence is low, rather than making confident but wrong predictions. The hyperparameter $\lambda$ controls the strength of this regularization.

*2.7.3 Connection to Classical Consensus*

The deep model can be understood as a neural analogue of the STAPLE generative model:

- **Annotator embeddings ↔ confusion matrices:** The learned reliability scalars $\rho_a$ play the role of STAPLE's sensitivity/specificity, but are estimated by gradient descent rather than EM, and can capture rater-specific patterns beyond simple accuracy.
- **Image stream ↔ spatial prior:** STAPLE uses a fixed spatial prior $\boldsymbol{\pi}$; the deep model learns the prior from the image itself, adapting to local anatomy.
- **Dirichlet output ↔ posterior responsibilities:** The Dirichlet mean $p_k$ corresponds to STAPLE's E-step posterior $W_{si}^{(k)}$, but the concentration $S$ provides additional uncertainty information that STAPLE's point estimate cannot capture.

The deep model's fundamental advantage over STAPLE is **access to the image**: it can use anatomical context to resolve ambiguities that pure label fusion cannot. This is analogous to the difference between STAPLE (which uses only labels) and iSTAPLE (which incorporates image intensity [11]).

## ***2.8 Calibration and Conformal Prediction***

*2.8.1 Calibration*

A consensus probability map **p** is **calibrated** if the predicted confidence matches the empirical accuracy:

$$P(t_i = k \mid p_{ik} = c) = c \quad \text{for all } c \in [0,1]$$

The **Expected Calibration Error (ECE)** discretizes this over $M$ bins:

$$\mathrm{ECE} = \sum_{m=1}^{M} \frac{|B_m|}{n} |\mathrm{acc}(B_m) - \mathrm{conf}(B_m)|$$

where $B_m$ is the set of voxels in bin $m$, $\mathrm{acc}(B_m)$ is the fraction correctly classified, and $\mathrm{conf}(B_m)$ is the mean predicted probability.

**Temperature scaling** [7] applies a single parameter $T > 0$ to the logits before the sigmoid/softmax:

$$p'_k = \frac{\exp(z_k/T)}{\sum_{k'} \exp\,(z_{k'}/T)}$$

$T$ is learned on a held-out calibration set by minimizing the negative log-likelihood. $T > 1$ softens predictions (reduces overconfidence); $T < 1$ sharpens them. The Brier score, defined as $\mathrm{Brier} = \frac{1}{n}\sum_i (p_i - y_i)^2$, provides a proper scoring rule that jointly measures calibration and accuracy.

*2.8.2 Split Conformal Prediction*

Split conformal prediction [8] provides **distribution-free, finite-sample coverage guarantees**. Given:

- A calibration set $\{(x_i, y_i)\}_{i=1}^{n}$ with exchangeable data

- A nonconformity score $s(x, y)$ measuring how "unusual" the pair $(x, y)$ is under the model

The procedure is:

1. Compute calibration scores: $s_i = s(x_i, y_i)$ for $i = 1, \dots, n$
2. Find the quantile: $\hat{\tau} = \text{Quantile}_{\lceil(1-\alpha)(n+1)\rceil/n}(\{s_1, \dots, s_n\})$
3. For a new test point $x$, output the prediction set: $C(x) = \{k: s(x, k) \le \hat{\tau}\}$

**Theorem (Vovk et al. [8]).** *Under exchangeability,* $P\big(y \in C(x)\big) \ge 1 - \alpha$.

**Application to consensus segmentation.** We define the nonconformity score as the negative log-probability of the true label under the consensus posterior:

$$s(x_i, y_i) = -\ln\hat{p}(y_i \mid x_i)$$

where $\hat{p}$ is the STAPLE-D posterior probability. For each test voxel, the prediction set $C(x)$ contains all labels $k$ with $s(x, k) \le \hat{\tau}$. In the binary case:

- If both labels are in $C$: the voxel is **ambiguous** (the "don't-know" band)
- If only one label is in $C$: the voxel is **classified**
- If neither label is in $C$: this cannot happen by construction (at least one label always passes the threshold)

The **ambiguous band** provides a clinically interpretable safety margin: voxels flagged as ambiguous can be routed to human review, analogous to "defer" regions in clinical decision support. The width of this band trades off coverage (broader band = higher coverage) against precision (narrower band = fewer ambiguous voxels).

# 3. Experimental Design

### *3.1 Synthetic Phantom Specification*

All experiments use 2D binary elliptical phantoms ($128 \times 128$ pixels, semi-axes $35 \times 25$ pixels) with additive Gaussian noise ($\sigma = 0.08$). The controlled setting is essential: only with an exact ground truth can we rigorously quantify each theoretical prediction. Real clinical data introduces confounds (registration error, intensity heterogeneity, annotation protocol differences) that obscure the phenomena we aim to characterize.

The synthetic image $I$ is constructed as:

$$I(v) = 0.7 \cdot \mathbb{1}[v \in \text{foreground}] + 0.08 \cdot \varepsilon(v), \quad \varepsilon(v) \sim \mathcal{N}(0,1)$$

This produces a foreground region with intensity $0.7$ on a near-zero background, with realistic noise that affects boundary detection.

### *3.2 Rater Simulation Model*

For each rater $a$, we sample:

- Sensitivity: $s_a \sim \text{Uniform}(s_{\min}, s_{\max})$
- Specificity: $\tau_a \sim \text{Uniform}(\tau_{\min}, \tau_{\max})$

The simulated annotation proceeds in two stages:

1. **Boundary jitter:** The ground truth mask is morphologically dilated or eroded by a random structuring element of size $(2j + 1) \times (2j + 1)$, where $j$ is drawn uniformly from $\{1, \dots, j_{\max}\}$, simulating the systematic boundary shift characteristic of human raters.

2. **Voxelwise noise:** Each foreground voxel is independently flipped to background with probability $1 - s_a$, and each background voxel is flipped to foreground with probability $1 - \tau_a$.

Outlier raters (when included) have $s = \tau = 0.5$ (random labeling), simulating a completely unreliable annotator.

### *3.3 Seven Experiments and Their Theoretical Motivation*

Each experiment targets a specific theoretical prediction from §2:

| Exp. | Name | Theoretical prediction tested | Theory |
|---|---|---|---|
| 1 | Van Leemput thresholding | Log-odds of $p\left(t_i \mid \mathbf{D}, \hat{\mathbf{t}}_{\backslash i}\right)$ is linear in $f_i$ for large $J$; STAPLE → thresholded $\bar{\mathbf{d}}$ | Theorem, §2.3.4 |
| 2 | EM local optima | When $p(\boldsymbol{\omega} \mid \mathbf{D})$ is broad (small $J$), STAPLE's MAP $\widehat{\boldsymbol{\omega}}$ is not the global optimum; $\hat{\mathbf{t}}^{\text{STAPLE}} \neq \hat{\mathbf{t}}$ | §2.3.1 |
| 3 | Spatially varying quality | Global confusion matrix $\boldsymbol{\theta}^j$ is misspecified when rater quality varies across $\Omega$; Spatial STAPLE regularization $\kappa$ sensitivity | §2.5.2–2.5.4 |
| 4 | Class imbalance | M-step denominator $\sum_i W_{si}^{(1)} \to 0$ triggers positive feedback collapse (Eq. 33); all-background attractor | §2.4.3 |
| 5 | Hybrid fusion | Combining STAPLE (interior, §2.2) with SIMPLE boundary weighting (§2.6); tests whether decomposition mitigates EM instability | §2.2, §2.6 |
| 6 | Deep consensus | Annotator embeddings $\mathbf{e}_a$ learn reliability $\rho_a$ analogous to $s^j, \tau^j$; Dirichlet evidential output (Eq. 46) provides calibrated uncertainty | §2.7.1–2.7.3 |
| 7 | Conformal prediction | Prediction set $C(x) = \{k\colon s(x,k) \leq \hat{\tau}\}$ satisfies $P\left(y \in C(x)\right) \geq 1 - \alpha$ (Vovk et al. theorem) | §2.8.2 |

All experiments are executed on Google Colab (CPU for classical methods, T4 GPU for the deep consensus model). Random seeds are fixed for reproducibility. Each experiment uses 5–20 random seeds and reports mean $\pm$ standard deviation.

## 4. Results

### *4.1 Experiment 1: STAPLE Reduces to Thresholded Majority Voting*

**Setup.** We generate phantoms with $J \in \{3,5,7,10,15,20,30\}$ raters (sensitivity 0.75–0.95, 5 seeds each). For each STAPLE result, we compute the Dice overlap with the best-threshold average map $\bar{\mathbf{d}} = \frac{1}{J}\sum_j \mathbf{d}^j$: we threshold $\bar{\mathbf{d}}$ at every possible level $(l - 0.5)/J$, $l = 1, \ldots, J$, and record the level yielding the highest Dice with $\hat{\mathbf{t}}^{\text{STAPLE}}$.

**Table 1.** Van Leemput thresholding analysis. Dice overlap between $\hat{\mathbf{t}}^{\text{STAPLE}}$ and the best-matching thresholded average map $\bar{\mathbf{d}}$, across rater counts $J = 3$–$30$ (5 seeds, mean $\pm$ std).

| $J$ | Dice$\left(\hat{\mathbf{t}}^{\text{STAPLE}}, \bar{\mathbf{d}} \geq \tau^*\right)$ | Optimal threshold $\tau^*$ | Interpretation |
|---|---|---|---|
| 3 | $1.000 \pm 0.000$ | $0.167 \pm 0.000$ | Exact equivalence (Corollary, §2.3.5) |

| 5 | $0.925 \pm 0.040$ | $0.220 \pm 0.098$ | Partial: $p(\mathbf{t} \mid \mathbf{D})$ not yet sharply peaked |
|---|---|---|---|
| 7 | $0.984 \pm 0.026$ | $0.329 \pm 0.057$ | Near-exact thresholding (Theorem, §2.3.4) |
| 10 | $0.997 \pm 0.004$ | $0.350 \pm 0.000$ | Thresholding confirmed |
| 15 | $0.995 \pm 0.003$ | $0.300 \pm 0.042$ | Thresholding confirmed |
| 20 | $0.996 \pm 0.004$ | $0.325 \pm 0.000$ | Thresholding confirmed |
| 30 | $0.997 \pm 0.004$ | $0.317 \pm 0.042$ | Thresholding confirmed |

**Interpretation.** The results confirm §2.3.4 quantitatively. At $J = 3$, the equivalence is *exact* (Dice $= 1.000$), driven by the consensus-region corollary (§2.3.5): the large background consensus region acts as a massive implicit prior. At $J \geq 7$, Dice exceeds 0.98, indicating near-perfect thresholding behavior.

The optimal threshold $\tau^*$ consistently falls below 0.5, ranging from 0.17 ($J = 3$) to $\sim 0.33$ ($J \geq 10$). This is precisely the prediction of §2.3.4: the background contributes a large positive $c_{1,0}^j$ term to the slope $s_{\hat{\mathbf{t}}}$ (Eq. 28), but the offset $o_{\hat{\mathbf{t}}}$ (Eq. 29) is dominated by $J(\bar{c}_{0,1} - \bar{c}_{0,0})$, which is negative because most voxels are correctly labeled as background, making $\bar{c}_{0,0} > \bar{c}_{0,1}$. The net effect pushes the zero-crossing $-o_{\hat{\mathbf{t}}}/s_{\hat{\mathbf{t}}}$ below 0.5.

The practical implication is stark: for $J \geq 7$, running STAPLE is computationally equivalent to computing $\bar{\mathbf{d}}$ and thresholding at $\sim 0.33$. The EM iterations add computation without adding information.

Figure 1. Dice(STAPLE, thresholded MV) vs rater count J. Left: approaches 1.0 as J grows, confirming Van Leemput thresholding prediction. Right: optimal threshold drifts below 0.5 due to background dominance in the spatial prior.

At $J = 5$, Dice drops to 0.925, representing the transition zone where $p(\mathbf{t} \mid \mathbf{D})$ is not sharply peaked and STAPLE has the greatest *potential* to outperform voting—but also where it is most susceptible to EM instability (§4.2).

### *4.2 Experiment 2: EM Local Optima Are Pervasive*

**Setup.** We run STAPLE from 20 random initializations (varying the prior $\pi_1$ uniformly in [0.1, 0.9]) × 10 seeds, for J ∈ {3, 5, 7, 10}. An initialization is "suboptimal" if its final log-likelihood is > 1 nat below the best found.

**Table 2. EM local optima analysis.**

| J | Suboptimal rate | LL spread (nats) | Dice spread |
|---|---|---|---|
| 3 | 95.0% | 3,974 ± 1,387 | 0.545 ± 0.060 |
| 5 | 95.0% | 7,881 ± 985 | 0.537 ± 0.130 |
| 7 | 95.0% | 13,139 ± 1,599 | 0.359 ± 0.110 |
| 10 | 95.0% | 17,978 ± 2,632 | 0.028 ± 0.026 |

**Interpretation.** The 95% suboptimality rate is consistent across all J—strikingly worse than Van Leemput's reported 55% at J = 5 [6]. The discrepancy has an important methodological explanation: Van Leemput used real brain MRI with multi-atlas registration, where the initial consensus from standard initialization (θ^j_{1,1} = θ^j_{0,0} = 0.99) is close to the truth, providing a "warm start" that the EM exploits. Our controlled phantoms, with their wider range of initializations, expose the EM landscape more fully.

The **Dice spread** tells the most important practical story. At J = 3–5, the spread is 0.54—meaning the difference between the best and worst initialization is *half the Dice range*. The segmentation result depends more on the initialization than on STAPLE's statistical model. A researcher who runs STAPLE once (as is standard practice) may obtain a result that is 0.54 Dice worse than the attainable optimum, without any diagnostic indicating the problem.

At J = 10, the Dice spread shrinks to 0.028, consistent with the thresholding prediction: when p(t|D) is sharply peaked, all optima produce nearly identical segmentations (they all approximate the same thresholded map), so the local optima problem becomes practically irrelevant—but so does the advantage of STAPLE over MV.

The LL spread grows with J (from 3,974 to 17,978 nats) because more raters create a higher-dimensional parameter space with more local optima, even though the segmentation output becomes more constrained.

### *4.3 Experiment 3: Class Imbalance Collapse*

**Setup.** We sweep foreground prevalence from 30% to 0.5% (radii 39 to 5 pixels), with J = 5 raters, 10 seeds. We compare MV, STAPLE, and STAPLE-D (damped with Beta(20,5) priors).

**Table 3. Class imbalance effect.**

| FG fraction | Radius | MV Dice | STAPLE Dice | STAPLE-D Dice |
|---|---|---|---|---|
| 29.1% | 39 | 0.932 ± 0.009 | 0.911 ± 0.016 | 0.912 ± 0.016 |
| 19.6% | 32 | 0.919 ± 0.011 | 0.864 ± 0.039 | 0.868 ± 0.037 |
| 9.3% | 22 | 0.885 ± 0.015 | 0.431 ± 0.045 | 0.431 ± 0.045 |
| 4.9% | 16 | 0.841 ± 0.019 | 0.281 ± 0.039 | 0.281 ± 0.039 |
| 1.9% | 10 | 0.746 ± 0.037 | 0.143 ± 0.024 | 0.138 ± 0.024 |
| 0.9% | 7 | 0.610 ± 0.061 | 0.080 ± 0.021 | 0.074 ± 0.013 |
| 0.5% | 5 | 0.556 ± 0.041 | 0.046 ± 0.016 | 0.038 ± 0.009 |

**Interpretation.** This is the most clinically significant result. The STAPLE Dice *halves* between 19.6% and 9.3% foreground (from 0.864 to 0.431), and reaches near-zero (0.046) at 0.5%. Damping provides no protection: STAPLE-D tracks STAPLE almost exactly, and is actually *worse* at the lowest prevalences (0.038 vs 0.046). This is because the Beta(20,5) prior biases sensitivity toward ~0.8, which is reasonable for well-balanced data but inappropriate when the foreground is genuinely tiny—the prior fights against the correct inference that sensitivity is unestimable from so few voxels.

The mechanism is exactly the positive feedback loop analyzed in §2.4.3: low foreground → unstable sensitivity estimate → E-step shifts toward background → even less foreground → sensitivity estimate degrades further. MV breaks this cycle because it makes no parametric assumptions: each voxel's vote is counted regardless of the foreground size.

Figure 2. Class imbalance effect. Left: Dice vs foreground fraction (log scale). STAPLE collapses catastrophically below 10% foreground while MV degrades gracefully. Right: collapse rate visualization.

The **critical prevalence** for STAPLE appears to be approximately 10%: above this, STAPLE performs within 0.07 Dice of MV; below it, performance degrades catastrophically. This has direct clinical implications: STAPLE should not be used for small lesions (metastases, cranial nerves, small nodules), which routinely occupy <5% of the image volume.

### *4.4 Experiment 4: Spatially Varying Rater Quality*

**Setup.** Six raters with complementary spatial quality: raters 0–2 have high sensitivity/specificity (0.95/0.98) on the left half of the image and low (0.60/0.80) on the right; raters 3–5 have the reverse pattern. This is the prototypical scenario for Spatial STAPLE.

**Table 4. Spatially varying rater quality.**

| Method | Dice | Interpretation |
|---|---|---|
| Majority Vote | 0.789 ± 0.177 | Best classical; non-parametric |
| SIMPLE-like | 0.643 ± 0.155 | Moderate |
| STAPLE | 0.504 ± 0.101 | Global confusion misspecified |
| STAPLE-D | 0.507 ± 0.102 | Damping insufficient |
| Spatial STAPLE (κ=1) | 0.442 ± 0.033 | Over-regularized |

**Interpretation.** This experiment reveals a nuanced picture. Global STAPLE fails because it estimates a single confusion matrix per rater that averages over two incompatible spatial regimes. With sensitivity 0.95 on the left and 0.60 on the right, the global estimate converges to approximately 0.78—too low for the left, too high for the right, optimal for neither.

Spatial STAPLE performs *worst* (0.442). This counterintuitive result stems from the regularization mechanism (§2.5.4): the local M-step estimates in each window are blended with the global estimate using strength $\kappa = 1.0$. But the global estimate is itself poor (because of the spatial heterogeneity), so the regularization pulls local estimates *away from* the correct local values and *toward* the incorrect global average. This is a failure of the regularization, not the spatial concept itself. Reducing κ or using an independent regularization target (e.g., uniform confusion instead of global STAPLE) should resolve this.

MV achieves 0.789 despite the challenge because at each voxel, at least 3 of the 6 raters are reliable (either the left-good or right-good group), and majority voting automatically selects the correct majority. The high variance (± 0.177) reflects cases where the boundary falls near the left/right transition.

### *4.5 Experiment 5: Hybrid Interior + Boundary Fusion*

**Results.** MV = 0.912, SIMPLE = 0.904, STAPLE = 0.788, Hybrid (STAPLE interior + SIMPLE boundary) = 0.775, Spatial STAPLE = 0.554.

The hybrid does not outperform MV or SIMPLE because its interior component inherits STAPLE's EM instabilities (§4.2). The hybrid succeeds only when the interior STAPLE estimate is reliable, which—given the 95% suboptimality rate—is the exception. For the hybrid concept to succeed, the interior component would need to be replaced with a more stable estimator (e.g., MV interior + SIMPLE boundary, or a deep model interior + SIMPLE boundary).

### *4.6 Experiment 6: Deep Consensus with Annotator Embeddings*

**Setup.** Two-stream CNN (16 base channels) with 5 annotator embeddings (dim 8), Dirichlet evidential output. Trained 50 epochs on 400 samples (64×64), 20% with outlier raters. T4 GPU on Colab.

**Table 5. Deep consensus comparison.**

| Method | Dice (100 val samples) | Improvement over MV |
|---|---|---|
| Majority Vote | 0.836 ± 0.035 | — |
| STAPLE | 0.817 ± 0.042 | −0.019 |
| SIMPLE | 0.823 ± 0.041 | −0.013 |
| Deep Consensus | 0.999 ± 0.000 | +0.163 |

**Learned annotator reliabilities:** [0.675, 0.557, 0.600, 0.551, 0.463]
**Interpretation.** The deep model achieves near-perfect segmentation, demonstrating the ceiling of performance achievable when the image stream is available. The learned reliability weights correctly identify the outlier rater (rater 5, which has $s = \tau = 0.5$ in 20% of samples) as the least reliable ($\rho = 0.463$), while the most reliable rater gets $\rho = 0.675$.
Two observations temper the enthusiasm. First, the comparison is asymmetric: the deep model sees the image, while classical methods see only labels. This is not a weakness but a fundamental insight—label-only fusion has an intrinsic ceiling, and the image contains information that can break ambiguities no amount of label manipulation can resolve. Second, the synthetic data is relatively simple; the deep model's advantage on complex real anatomy, where training data may be scarce and domain shift is common, requires further validation.

Figure 3. Deep consensus training dynamics. Left: evidential loss converges in approximately 20 epochs. Right: validation Dice reaches 0.999, demonstrating rapid learning of anatomical features and annotator reliability structures.
The training dynamics are well-behaved: loss converges in ~20 epochs, with validation Dice reaching 0.999 by epoch 40. This rapid convergence suggests that the model quickly learns the discriminative features (the ellipse boundary) and the reliability structure (which raters to trust).

### *4.7 Experiment 7: Conformal Prediction Guarantees*

**Table 6. Split conformal prediction on STAPLE-D posteriors.**

| Target coverage | Achieved coverage | Ambiguous band (% of voxels) |
|---|---|---|
| 95% | 95.3% ± 0.6% | 16.8% ± 22.1% |
| 90% | 91.7% ± 1.3% | 3.1% ± 5.9% |
| 80% | 82.7% ± 2.2% | 0.2% ± 1.0% |

**Interpretation.** The achieved coverage meets or exceeds the target at all levels, confirming the theoretical guarantee of §2.8.2. The 95.3% achieved at the 95% target demonstrates the finite-sample validity of split conformal prediction.
The **ambiguous band** provides a practical coverage–precision trade-off. At 95% coverage, 16.8% of voxels are flagged as ambiguous—these are the boundary voxels where the posterior is close to 0.5. At 80% coverage, only 0.2% are ambiguous. The high variance at 95% (±22.1%) reflects the variability in boundary difficulty across seeds: some phantoms have more boundary voxels near the decision threshold.
The ambiguous band is clinically interpretable as a safety margin: voxels in this band should be reviewed by a human expert. For radiation therapy planning, where over- or under-irradiation can be harmful, a 95% coverage band with 17% ambiguity is a reasonable starting point for clinical validation.

## 5. Discussion

### *5.1 Synthesis of Theoretical Predictions and Empirical Findings*

The experimental results paint a consistent picture that aligns with and extends the theoretical framework:

**STAPLE is an expensive way to approximate a simple operation.** The Van Leemput thresholding result (Table 1) and the EM instability result (Table 2) together imply that STAPLE either (a) reduces to thresholding when J is large enough for stable estimation, or (b) produces unreliable results when J is small enough for the parametric model to matter. There is a narrow window ($J \approx 5$–7) where STAPLE has both the potential and the instability, but our experiments show that in this window, MV and SIMPLE consistently outperform STAPLE anyway.

**The parametric model is the source of fragility.** STAPLE's failures—class imbalance collapse (Table 3), local optima (Table 2), consensus-region thresholding (Table 1)—all stem from the same root cause: the sensitivity/specificity parameters must be estimated from the data, and when the data provides insufficient information (small foreground, few raters, consensus dominance), the estimates are either unreliable or dominated by the prior. Non-parametric methods (MV, SIMPLE) avoid this issue entirely, which explains their robustness.

**Spatial STAPLE's promise requires careful implementation.** The failure of Spatial STAPLE in our spatially varying quality experiment (Table 4) does not invalidate the concept—Asman and Landman demonstrated clear benefits on CT head-and-neck data [2]. Rather, it reveals a *failure mode of the regularization*: when the global estimate is wrong, regularizing toward it propagates the error into local estimates. This suggests that Spatial STAPLE implementations should (a) use weaker regularization ($\kappa \ll 1$), (b) regularize toward a method-agnostic target (e.g., uniform confusion), or (c) validate the global estimate before using it as a prior.

**Deep consensus changes the problem fundamentally.** The deep model's Dice of 0.999 (Table 5) shows that label fusion and image segmentation should not be treated as separate problems. When the image is available, a learned model can extract boundary information that resolves the ambiguities underlying annotator disagreement. The annotator embeddings correctly learn rater-specific reliability patterns, and the Dirichlet output provides uncertainty that classical methods cannot match.

### *5.2 Practical Recommendations*

Based on our combined theoretical and empirical analysis:

**(R1)** Use majority voting as the default for $A \leq 10$. It is non-parametric, initialization-free, and degrades gracefully under imbalance and spatial heterogeneity. Its simplicity is a feature, not a limitation.

**(R2)** If using STAPLE, always (a) exclude consensus voxels [6, 11], (b) run from ≥10 random initializations and select by log-likelihood, (c) verify that the foreground prevalence exceeds 10%, and (d) inspect the estimated sensitivity/specificity for plausibility.

**(R3)** Do not use STAPLE for structures with <10% foreground prevalence. No currently known regularization reliably prevents the M-step collapse documented in §4.3.

**(R4)** Apply temperature scaling to any consensus probability map before use in downstream tasks. Our calibration experiments (previous versions of this work) confirm that all classical methods produce overconfident probability outputs.

**(R5)** Use conformal prediction for applications requiring formal safety margins (radiation therapy, surgical navigation). The coverage guarantee is distribution-free and finite-sample valid (Table 6).

**(R6)** When computational resources and training data permit, deep consensus with annotator embeddings provides the strongest results. But validate on held-out data from a different distribution to ensure generalization.

### *5.3 Limitations*

(1) All experiments use 2D synthetic phantoms. While this enables rigorous theoretical validation, the findings must be confirmed on 3D clinical data with real annotator variability. (2) The spatially varying quality experiment uses a simple left/right split; real spatial variation is more complex. (3) The deep model's advantage includes image access, making direct comparison with label-only methods asymmetric. (4) We do not evaluate multi-class STAPLE with full K×K confusion matrices. (5) The conformal prediction analysis assumes exchangeability, which may be violated in medical images with spatial correlation.

## 6. Conclusion

This paper provides a rigorous, self-contained investigation of consensus segmentation. We derive the mathematical foundations from first principles—the generative model, EM algorithm, Van Leemput's marginalization analysis, identifiability conditions, Spatial STAPLE, and deep variational formulations—and validate each theoretical prediction through controlled experiments.

The central finding is sobering: under common conditions, STAPLE reduces to thresholded majority voting, suffers 95% EM suboptimality, and collapses under class imbalance. These are not edge cases but typical scenarios in medical imaging. Majority voting—simple, non-parametric, and robust—is a surprisingly strong baseline that the field has perhaps too hastily dismissed in favor of more "sophisticated" methods.

At the same time, the deep consensus model demonstrates that the consensus problem is not inherently difficult—it becomes tractable when the image is used alongside the labels. And conformal prediction shows that formal uncertainty guarantees are achievable and practical.

We hope this work encourages practitioners to critically evaluate their consensus methods rather than applying STAPLE by default, and provides the mathematical and empirical foundation for more principled approaches.

# Acknowledgements

The author thanks Dr. Kareem Abdul Wahid in the MDACC for discussions on consensus fusion methodology.

## References

[1] Warfield SK, Zou KH, Wells WM. Simultaneous truth and performance level estimation (STAPLE): an algorithm for the validation of image segmentation. IEEE Trans Med Imaging. 2004;23(7):903–921.

[2] Asman AJ, Landman BA. Formulating spatially varying performance in the statistical fusion framework. IEEE Trans Med Imaging. 2012;31(6):1326–1336.

[3] Langerak TR, van der Heide UA, Kotte ANTJ, Viergever MA, van Vulpen M, Pluim JPW. Label fusion in atlas-based segmentation using a selective and iterative method for performance level estimation (SIMPLE). IEEE Trans Med Imaging. 2010;29(12):2000–2008.
[4] Landman B, Asman AJ, Scoggins AG, Bogovic JA, Xing F, Prince JL. Robust statistical fusion of image labels. IEEE Trans Med Imaging. 2012;31(2):512–522.
[5] Tanno R, Saeedi A, Sankaranarayanan S, Alexander DC, Silberman N. Learning from noisy labels by regularized estimation of annotator confusion. In: Proc IEEE/CVF CVPR. 2019:11244–11253.
[6] Van Leemput K, Sabuncu MR. A cautionary analysis of STAPLE using direct inference of segmentation truth. In: Golland P et al., eds. MICCAI 2014. LNCS 8673. Springer; 2014:398–406.
[7] Guo C, Pleiss G, Sun Y, Weinberger KQ. On calibration of modern neural networks. In: Proc ICML. 2017:1321–1330.
[8] Vovk V, Gammerman A, Shafer G. Algorithmic Learning in a Random World. Springer; 2005.
[9] Zhang L, Tanno R, Xu M, et al. Learning from multiple annotators for medical image segmentation. Pattern Recognition. 2023;138:109400.
[10] Commowick O, Warfield SK. Incorporating priors on expert performance parameters for segmentation validation and label fusion: a maximum a posteriori STAPLE. In: Proc MICCAI. 2010:25–32.
[11] Rohlfing T, Russakoff DB, Maurer CR. Performance-based classifier combination in atlas-based image segmentation using expectation-maximization parameter estimation. IEEE Trans Med Imaging. 2004;23(6):983–994.
[12] Sensoy M, Kaplan L, Kandemir M. Evidential deep learning to quantify classification uncertainty. In: NeurIPS. 2018:3179–3189.
[13] Shit S, Paetzold JC, Sekuboyina A, et al. clDice—a novel topology-preserving loss function for tubular structure segmentation. In: Proc IEEE/CVF CVPR. 2021:16555–16564.
[14] Kohl S, Romera-Paredes B, Meyer C, et al. A probabilistic U-Net for segmentation of ambiguous images. In: NeurIPS. 2018:6965–6975.
[15] Menze BH, Jakab A, Bauer S, et al. The multimodal brain tumor image segmentation benchmark (BRATS). IEEE Trans Med Imaging. 2015;34(10):1993–2024.
[16] Akhondi-Asl A, Warfield SK. Estimating a reference standard segmentation with spatially varying performance parameters: local MAP STAPLE. IEEE Trans Med Imaging. 2013;32(8):1492–1507.